\documentclass{article}
\usepackage{spconf,amsmath,epsfig}
\usepackage{url}
\usepackage{hhline}
\usepackage{array}

\title{Cloud-Net: An end-to-end Cloud Detection Algorithm for Landsat 8 Imagery}
\name{Sorour Mohajerani, Parvaneh Saeedi\thanks{Our thanks goes to the Government of Canada for providing financial support for this project through the Technology Development Program.}}
\address{School of Engineering Science, Simon Fraser University, Burnaby, BC, Canada}
\begin{document}
%
\maketitle
\begin{abstract}
Cloud detection in satellite images is an important first-step in many remote sensing applications. This problem is more challenging when only a limited number of spectral bands are available. To address this problem, a deep learning-based algorithm is proposed in this paper. This algorithm consists of a Fully Convolutional Network (FCN) that is trained by multiple patches of Landsat 8 images. This network, which is called Cloud-Net, is capable of capturing global and local cloud features in an image using its convolutional blocks. Since the proposed method is an end-to-end solution no complicated pre-processing step is required. Our experimental results prove that the proposed method outperforms the state-of-the-art method over a benchmark dataset by 8.7\% in Jaccard Index.
\end{abstract}
\vspace{-5mm}
\section{Introduction}
\label{sec:intro}

Precise identification/measurement of cloud coverage is a crucial step in the analysis of satellite imagery. For instance, clouds could occlude objects on the land and cause difficulty for many remote sensing applications including change detection, geophysical parameter retrieving, and object tracking \cite{change_detect,geophys,hurricane2}. In addition, transmission of images with high cloud coverage from satellites to the ground stations would be unnecessary and useless. Cloud coverage by itself might provide useful information about the climate parameters and natural disasters such as hurricanes and volcanic eruptions \cite{volcano, hurricane}. As a result, identification of the cloud regions in images is an important pre-processing step for many applications.

In recent years, researchers have developed many cloud detection methods. These methods can be divided into three main categories: threshold-based \cite{acca,fmask1,fmask2,fmask3}, handcrafted \cite{hot,bag}, and deep learning-based \cite{multilevel, mymmsp} approaches. Zhu et al. in \cite{fmask1} introduced the Function of mask (Fmask) algorithm. Fmask basically consisted of a decision tree in which the potential cloud pixels were separated from non-cloud pixels based on multiple threshold functions. An improved version of Fmask was proposed in \cite{fmask2}. This method benefited from Cirrus band of Landsat 8 to increase the accuracy of the detected clouds and is currently utilized to produce cloud masks of the Landsat Level-1 data products \cite{web}. Qui et al. in \cite{fmask3} integrated Digital Elevation Map (DEM) information into Fmask and improved its performance in mountainous areas.

Haze Optimized Transformation (HOT) \cite{hot} is among the most famous handcrafted algorithms for identification of clouds \cite{bag}. In this algorithm, Zhang et al. utilized the correlation between two spectral bands of Landsat images to distinguish thin clouds from clear regions.

In recent years, deep learning-based methods have been proved to deliver good performance in many image processing applications. Researchers, in the remote sensing field, have also proposed such algorithms to address the problem of cloud detection. For instance, Xie et al. \cite{multilevel} utilized two Convolutional Neural Networks (CNNs) for classification of sub-regions in an image into thick cloud, thin cloud, or non-cloud classes. Many Fully Convolutional Neural Networks (FCNs) have been introduced for semantic segmentation of images. Most of these networks have an encoder-decoder architecture, which is inspired by U-Net \cite{unet}. The effectiveness of U-Net has been proven in many other computer vision applications \cite{3DUNet,reza-unet}. Authors in \cite{mymmsp} used a FCN to segment the cloud regions of Landsat 8 images. This network was a very deep version of U-Net and was trained by images and their automatically generated Ground Truths (GTs).

In this paper, we introduce a new Cloud detection Network (Cloud-Net) for end-to-end pixel-level labeling of the satellite images. We specifically designed convolution blocks of Cloud-Net to capture complicated semantic features of clouds in remote sensing images. Unlike Fmask and the method in \cite{multilevel}, Cloud-Net is capable of learning both local and global features from the entire scene in an end-to-end manner. It does not require any complicated pre-processing step such as super-pixel segmentation. In addition, unlike \cite{mymmsp}, Cloud-Net effectively utilizes the extracted features from its sophisticated convolution blocks to recover more accurate cloud masks. As a result, it delivers superior performance for cloud detection. Moreover, we have modified the dataset introduced in \cite{mymmsp} since it included a few images with inaccurate and uncertain GTs. This new dataset is publicly available to the geospatial community by request.

\section{Methodology}
\label{sec:format}
\begin{figure}[t]
\begin{minipage}[b]{1.0\linewidth}
  \centering
  \centerline{\epsfig{figure=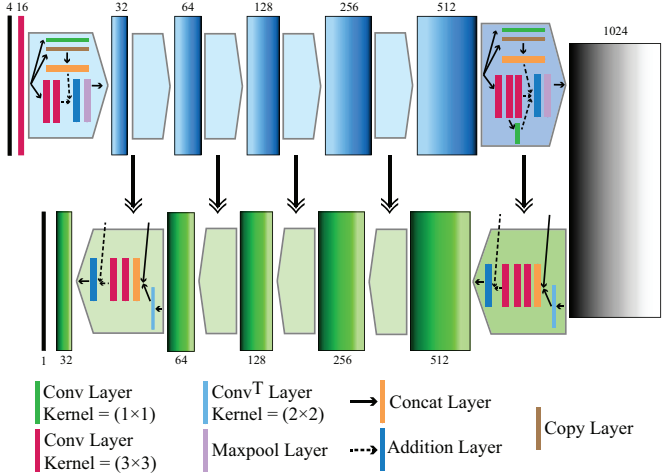,width=8.7cm}}
\end{minipage}
\caption{\footnotesize Cloud-Net architecture. Conv\textsuperscript{T}, Concat, and Maxpool refer to convolution transposed, concatenation, and maxpooling, respectively. The bars with gradient shading represent the feature maps. The numbers on the top and the bottom of the bars are the corresponding depth of each feature map.}
\label{fig:arch}
\vspace{-4mm}
\end{figure}

To get a cloud mask at a CNN's output with the same size as the input image, the CNN should have two branches or arms. One of these arms--contracting arm--is responsible for extracting features and producing deep low-level features of the input image. The other arm--expanding arm--is to utilize those features and retrieve cloud attributes, recover them, and finally generate an output cloud mask. This output is, in fact, a probability map in which every pixel location correspond to the probability of that pixel belonging to the cloud class. Our cloud segmentation CNN, Cloud-Net, shares the same contracting and expanding arms as mentioned above. Indeed, it is inspired by the network used in \cite{mymmsp_shadow}. The architecture of the proposed Cloud-Net is displayed in Fig. \ref{fig:arch}. The blue bars and blocks form the contracting arm, while the green arrows and blocks build the expanding arm.

We have used Landsat 8 spectral images to train Cloud-Net. Landsat 8 contains two optical sensors to acquire eleven spectral bands. In this paper, we have utilized four of these bands--Band 2 to Band 5--since they are among the more common bands provided by most remote sensing satellites such as Sentinel-2, HJ-1, GF-2, etc. Since the spatial dimensions of the Landsat 8 images are very large, multiple non-overlapping $384\times 384$ patches are extracted from each of those images. Before being utilized by Cloud-Net, these patches are downsized to $192\times 192$. Therefore input size of the network is $192\times 192\times 4$ and the size of the output cloud mask is $192\times 192\times 1$.

The shortcut connections in each block (consisting of Concat, Copy, and Addition layers shown in Fig. \ref{fig:arch}) help the network to preserve and utilize the learned contexts from the earlier layers. As a result, the network is capable of capturing more cloud features. Another effect of these connections is speeding up the training process by preventing the network from experiencing the vanishing gradient  phenomenon during backpropagation. In addition, the connections between two arms of the Cloud-Net help the expanding arm to generate a more accurate cloud mask.

Before each training epoch, commonly used geometric data augmentation techniques such as horizontal flipping, rotation, and zooming are applied. The activation function of the convolution layers of Cloud-Net is ReLU \cite{relu}. A sigmoid layer is used in the last convolution layer of the network. For optimization of our model, the Adam gradient descent method \cite{ADAM} is used. The following soft Jaccard loss function \cite{reza-unet,jacc1} is implemented to be minimized:

\vspace{-5mm}
\begin{equation}
\begin{split}
F_{L}(t,y) \! = \!-\dfrac{\sum\limits_{i=1}^{N} t_{i} y_i+\epsilon}{\sum\limits_{i=1}^{N} t_{i} + \sum\limits_{i=1}^{N} y_i - \sum\limits_{i=1}^{N} t_{i} y_i+\epsilon}\raisebox{0.7mm}{.}
\\ 
\end{split}
\label{Eq:loss}
\end{equation}
Here, $t$ denotes the GT and $y$ is the output array of Cloud-Net. $y_i$ and $t_{i}$ represent the $i$th pixel value of $y$ and $t$, respectively. $N$ is the total number of pixels in the GT. To avoid division by zero, $\epsilon = 10^{-7}$ is added to the numerator and denominator of the loss function.

The weights of the network are initialized with a uniform random distribution between $[-1, 1]$. The initial learning rate for the training of the model is set to $10^{-4}$. We applied a learning rate decay policy during the training phase with a decay rate of $0.7$ and a patience factor of $15$. This policy is continued until the learning rate reaches the value of $10^{-9}$. The proposed network is implemented using Keras framework.

\section{Experimental Setting}
\label{sec:pagestyle}

\subsubsection{Dataset}
For training and testing of the proposed method, we have utilized the dataset introduced in \cite{mymmsp} with a few modifications. This dataset has 18 Landsat 8 images for training and 20 images for testing. Among these 38 images, we have noticed that five of them (four from the training set and one from the test set) had inaccurate and uncertain GTs. We replaced these images with five new images. Next, the GTs of all images in the training set have been manually annotated. Therefore, instead of using automatically generated GTs of the training set in \cite{mymmsp}, we use more accurate and manually obtained GTs of the modified training set. This obviously affects the performance of the algorithm since the network learns the cloud features from correct images/GTs instead of inconsistent data. It is worth noting that after cropping the images of this dataset into $384\times384$ patches, 8400 patches for training and 9201 patches for testing are obtained. Cloud-Net is trained with 8400 training patches. This dataset, which is called 38-Cloud dataset, is publicly available to the research community by request.

\subsubsection{Test Phase}
To obtain the cloud mask of an unseen test image, it is first cut into multiple $384\times384$ non-overlapping patches. Then, each patch is resized to $192\time192$ (to be fit the input of the network) and fed into the Cloud-Net. The predicted cloud probability map of each patch is obtained using the well-trained weights of our network. Next, it is binarized using a global threshold of $0.047$ and then resized to $384\times384$. Once the predicted masks of all the patches are produced, they are stitched together to create the final cloud mask of the original test image.

\begin{figure}[htb]
\centering
\begin{minipage}{0.2\textwidth}
\centering
\centerline{\includegraphics[height=35mm, width=35mm]{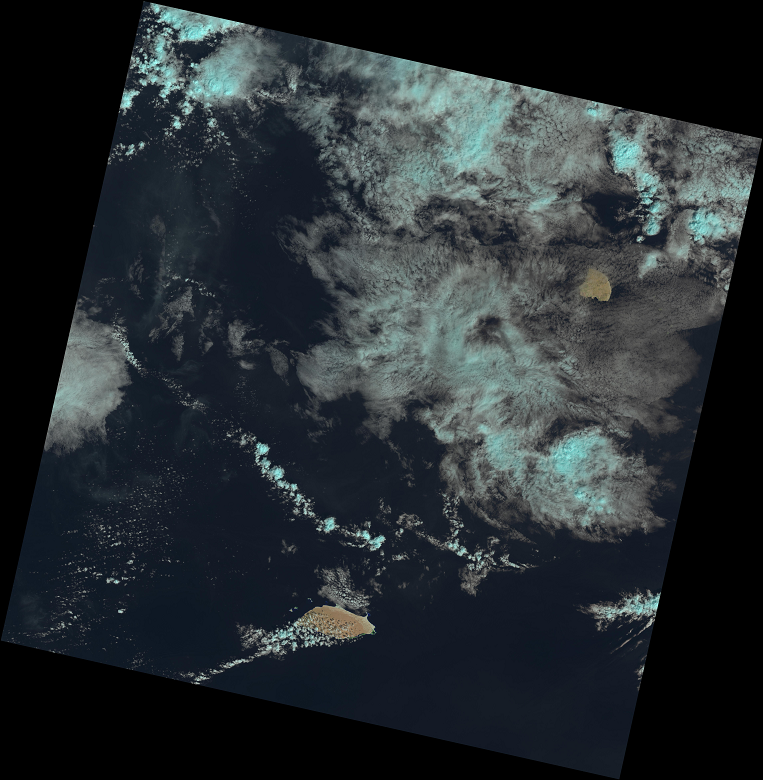}}\footnotesize{(a)}
\end{minipage}
\vspace{1mm}
\hspace{5mm}
\begin{minipage}{0.2\textwidth}
\centering
\centerline{\includegraphics[height=35mm, width=35mm]{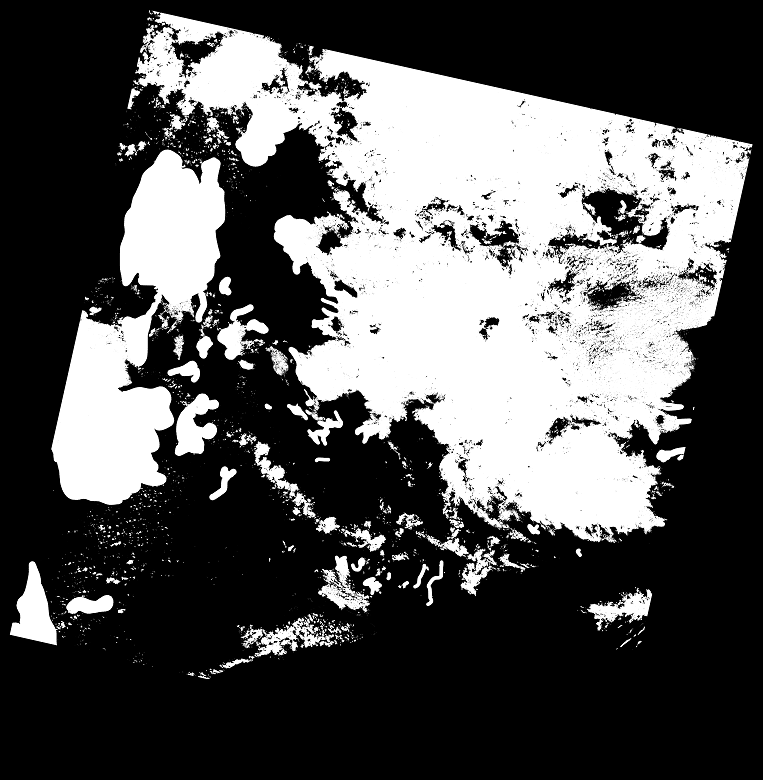}}\footnotesize{(b)}
\end{minipage}
\begin{minipage}{0.2\textwidth}
\centering
\centerline{\includegraphics[height=35mm, width=35mm]{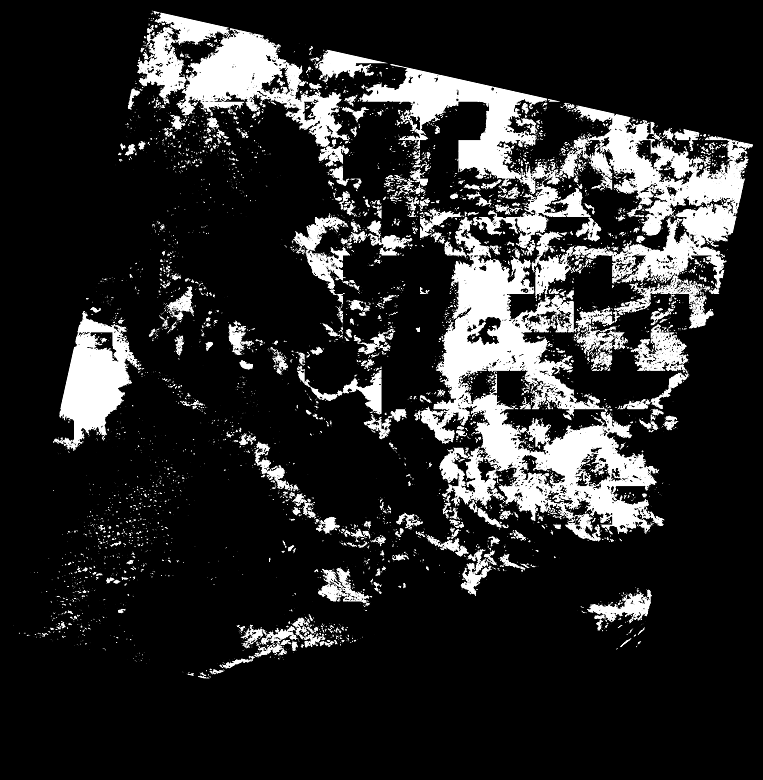}}
\footnotesize{(c)}
\end{minipage}
\vspace{1mm}
\hspace{5mm}
\begin{minipage}{0.2\textwidth}
\centering
\centerline{\includegraphics[height=35mm, width=35mm]{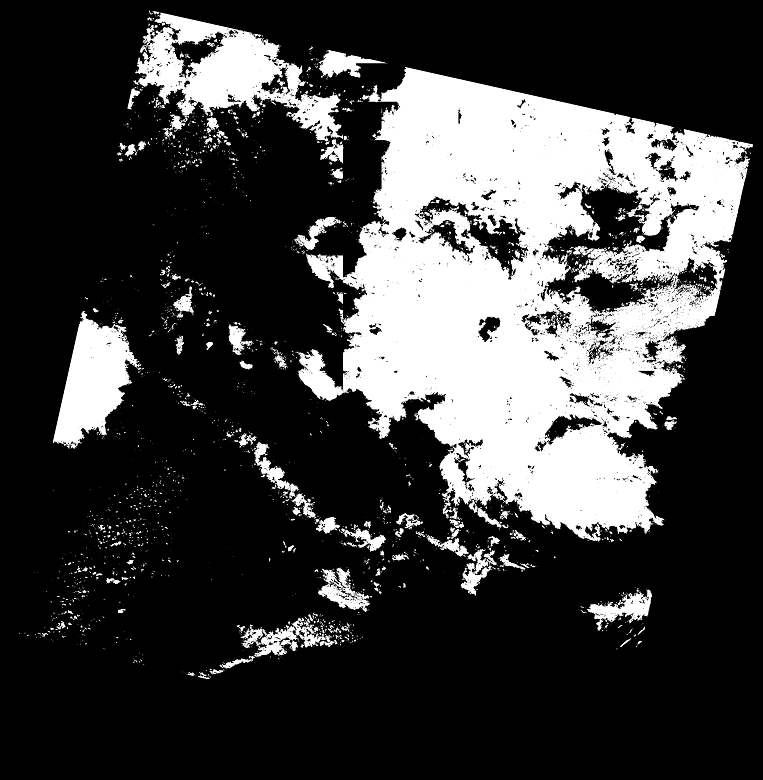}}
\footnotesize{(d)}
\end{minipage}
\begin{minipage}{0.2\textwidth}
\centering
\centerline{\includegraphics[height=35mm, width=35mm]{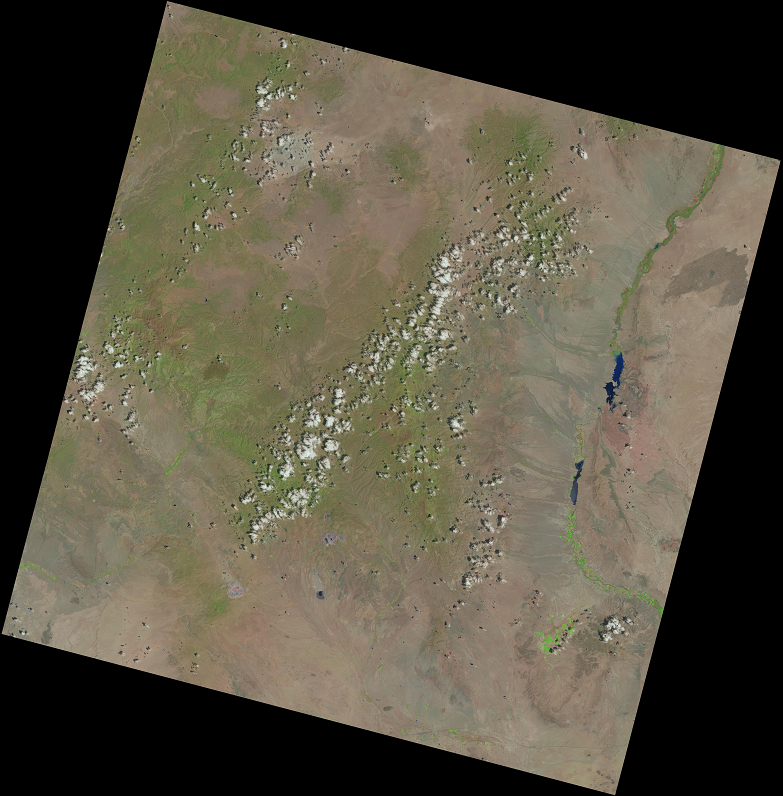}}\footnotesize{(e)}
\end{minipage}
\vspace{1mm}
\hspace{5mm}
\begin{minipage}{0.2\textwidth}
\centering
\centerline{\includegraphics[height=35mm, width=35mm]{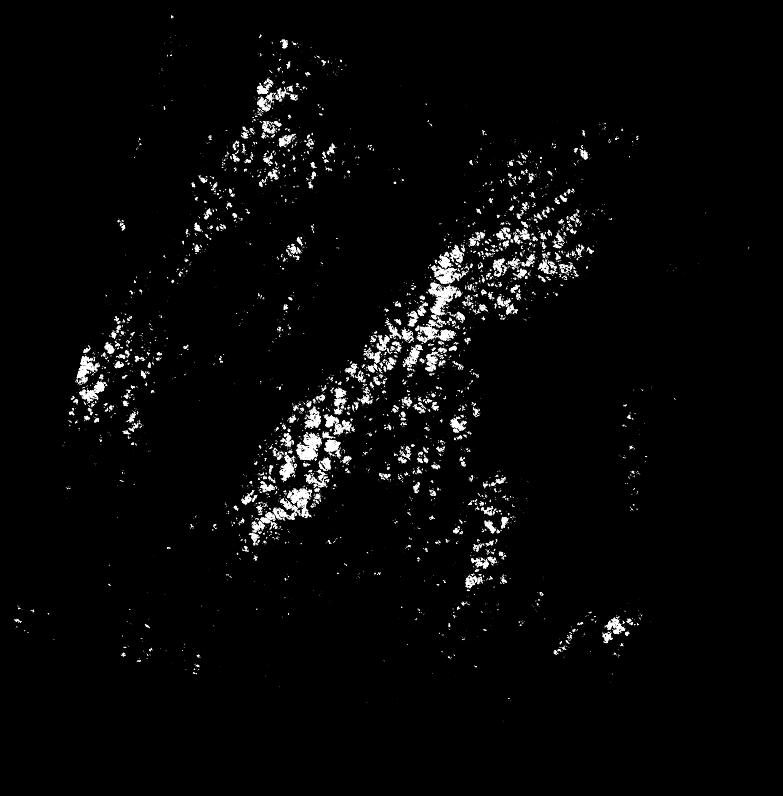}}\footnotesize{(f)}
\end{minipage}
\begin{minipage}{0.2\textwidth}
\centering
\centerline{\includegraphics[height=35mm, width=35mm]{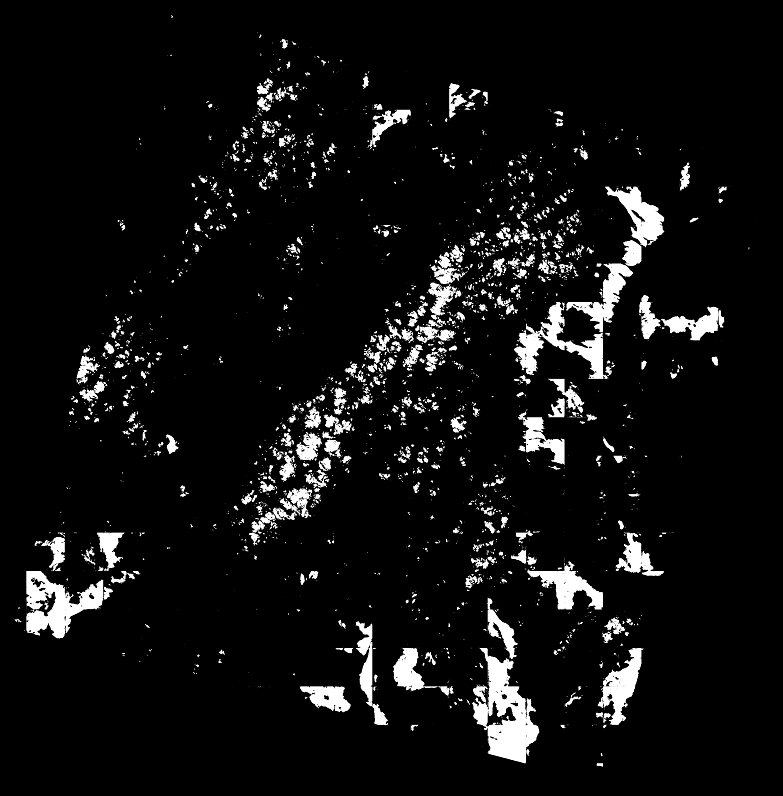}}
\footnotesize{(g)}
\end{minipage}
\vspace{1mm}
\hspace{5mm}
\begin{minipage}{0.2\textwidth}
\centering
\centerline{\includegraphics[height=35mm, width=35mm]{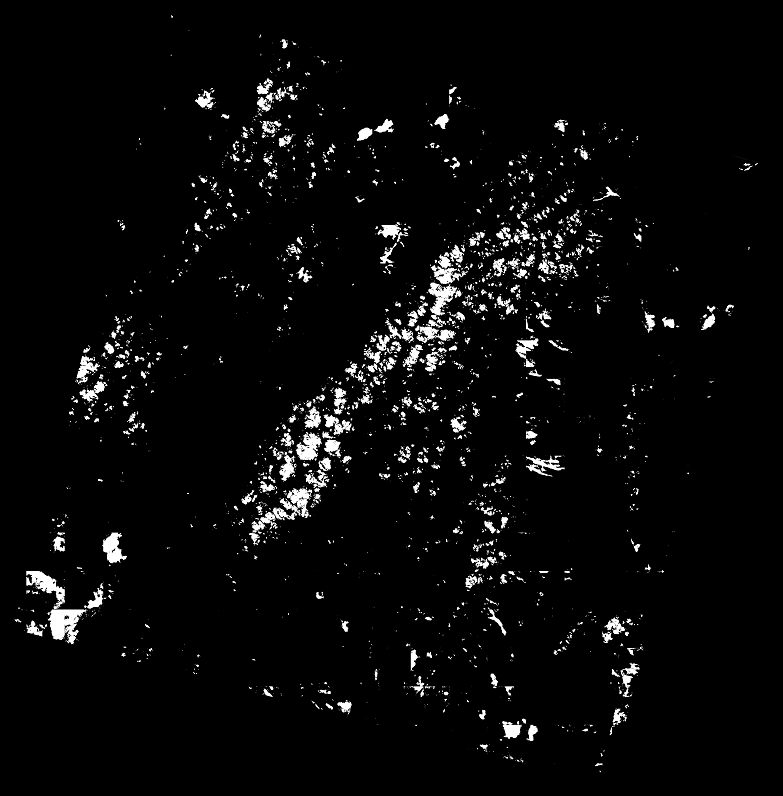}}
\footnotesize{(h)}
\end{minipage}
\hspace{-1mm}  
\setlength{\abovecaptionskip}{2mm}
\caption{\footnotesize Cloud-Net visual results: (a),(e) natural color images from 38-Cloud test set, (b),(f) corresponding GTs (c),(g) results of FCN \cite{mymmsp} (d),(h) results of Cloud-Net.\label{Fig:samples}}
\end{figure}

\subsubsection{Evaluation Metrics}
Once the cloud mask of a complete Landsat 8 scene is generated, it is compared with the corresponding GTs. The predicted mask has two classes of "cloud" and "clear" for each pixel. The performance of our algorithm is quantitatively measured by evaluating the Overall Accuracy, Recall, Precision, Specificity, and Jaccard Index. These metrics are defined as follows:

\vspace{-5mm}
\begin{equation}
\small
\begin{split}
&\quad \text{Jaccard Index}=\frac{TP}{TP+FN+FP}\raisebox{0.7mm}{,} \\
&\quad \text{Precision}=\frac{TP}{TP+FP}\raisebox{0.7mm}{,} \\
&\quad \text{Recall}=\frac{TP}{TP+FN}\raisebox{0.7mm}{,} \\
&\quad \text{Specificity}=\frac{TN}{TN+FP} \raisebox{0.7mm}{,}\\
&\quad \text{Overall Accuracy}=\frac{TP+TN}{TP+TN+FP+FN} \raisebox{0.7mm}{.}\\
\end{split}
\label{Eq:Ev1}
\end{equation}

Here TP, TN, FP, and FN are the total number of true positive, true negative, false positive, and false negative pixels, respectively. The Jaccard Index or intersection over union is a widely accepted metric for measuring the performance of many image segmentation algorithms \cite{mymmsp}.

\section{Experimental Results}
Table \ref{Tab:numerical} represents the quantitative results of the proposed method over 20 test images of the 38-Cloud dataset. As shown in this table, Cloud-Net improves FCN's Jaccard Index by 8.7\%. This indicates the superiority of our proposed network to FCN given that both are trained with the same training set. Indeed, to get the numerical results in the first row of Table \ref{Tab:numerical}, we have trained FCN with the exact setting as \cite{mymmsp} using 38-Cloud training set and their more accurate GTs. Therefore, the numerical results of this experiment are fair to be compared with that of Cloud-Net. Please note that according to \cite{mymmsp}, FCN's overall accuracy is 88.30\%. Also, the proposed method exceeds Fmask's performance, which is a widely used algorithm for cloud detection. Some qualitative results of the proposed method are displayed in Fig. \ref{Fig:samples}.

\renewcommand{\tabcolsep}{3pt}
\vspace{-4mm}
\begin{table}[h]
\small
\begin{minipage}[t]{0.49\textwidth}
\centering
\caption{Cloud-Net performance over 38-Cloud test set (in~\%).
\vspace{-2mm}
\label{Tab:numerical}} 
\begin{tabular}{|m{17mm}|c|c|c|c|m{13mm}|}
\hline
\centering \textbf{Method}   & \textbf{Jaccard}                       & \textbf{Precision}   & \textbf{Recall} & \textbf{Specificity} & \textbf{Overall Accuracy}   \\ 
\hhline{|=|=|=|=|=|=|}
FCN \cite{mymmsp} trained with 38-Cloud training set & 72.17 & 84.59
  & 81.37 &   98.45  & \hspace{2mm} 95.23 \\ \hline
Fmask \cite{fmask2}    & 75.16  & 77.71 & \textbf{97.22} &  93.96 & \hspace{2mm} 94.89 \\ \hline
Proposed Cloud-Net  & \textbf{78.50} &  \textbf{91.23}&  84.85  & \textbf{98.67} & \hspace{2mm} \textbf{96.48}
\\ \hline
\end{tabular}
\end{minipage}
\vspace{-5mm}
\end{table}

\vspace{-2mm}
\section{Conclusion}
In this paper, a deep learning-based algorithm for segmentation of clouds in remote sensing images was developed. The proposed FCN, Cloud-Net, benefits from sophisticated convolution blocks. Using a specialized architecture, Cloud-Net performance is superior to the competing state-of-the-art algorithms. In addition, we modified the dataset introduced in \cite{mymmsp}, giving researchers access to a more reliable and more accurate dataset for future analysis.

\vspace{-5mm}
\bibliographystyle{IEEEbib}
\bibliography{strings,refs}

\end{document}